\documentclass[10pt,twocolumn,letterpaper]{article}

\usepackage[pagenumbers]{cvpr} %

\usepackage{graphicx}
\usepackage{amsmath}
\usepackage{amssymb}
\usepackage{booktabs}

\usepackage{algorithm}
\usepackage{algpseudocode}

\usepackage[pagebackref,breaklinks,colorlinks]{hyperref}

\usepackage[capitalize]{cleveref}
\crefname{section}{Sec.}{Secs.}
\Crefname{section}{Section}{Sections}
\Crefname{table}{Table}{Tables}
\crefname{table}{Tab.}{Tabs.}

\begin{document}

\title{RoomDreamer: Text-Driven 3D Indoor Scene Synthesis with Coherent Geometry and Texture}

\author{
Liangchen Song$^{12}$, Liangliang Cao$^{1}$, Hongyu Xu$^{1}$\\ 
Kai Kang$^{1}$, Feng Tang$^{1}$, Junsong Yuan$^{2}$, Yang Zhao$^{1}$ \\
{
$^1$ Apple Inc. \quad $^2$ University at Buffalo 
}
\\
\small{
\texttt{lsong8@buffalo.edu \quad llcao@apple.com}
}
}

\twocolumn[{%
\renewcommand\twocolumn[1][]{#1}%
\maketitle
\begin{center}
    \centering
    \includegraphics[width=\textwidth]{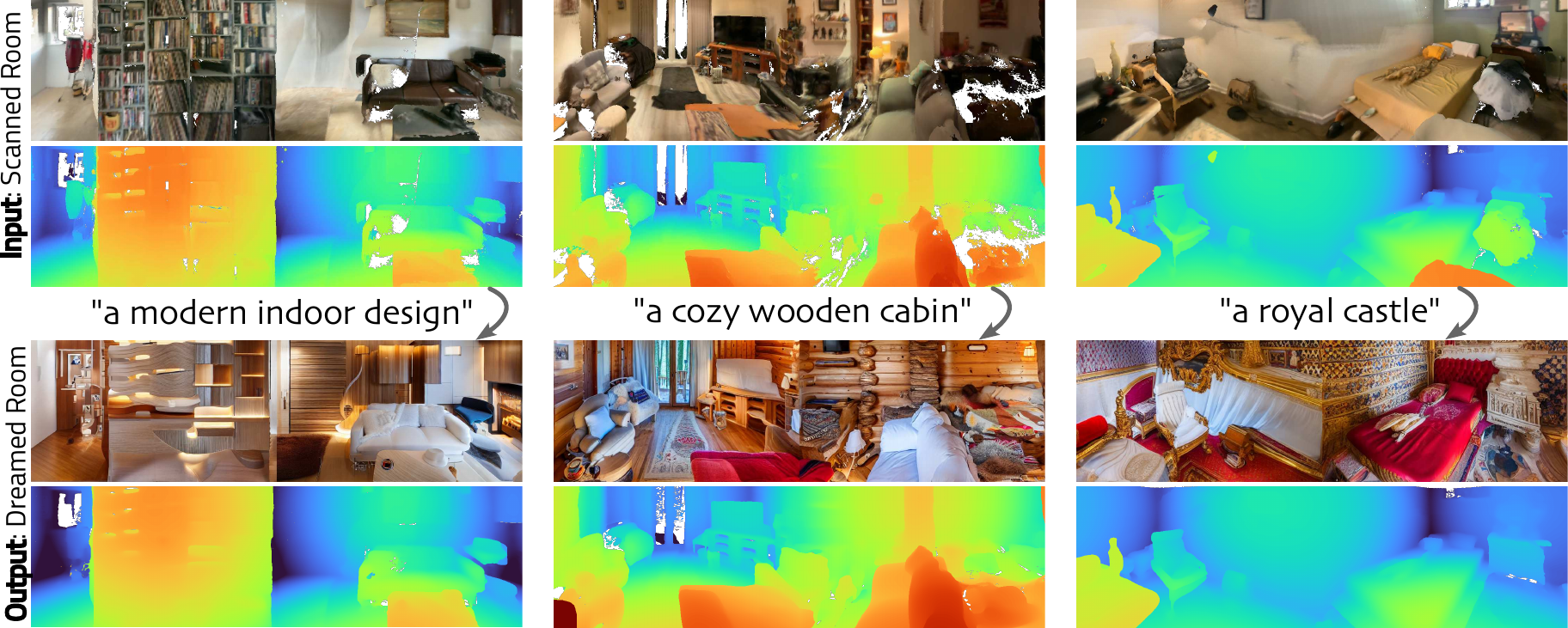}
    \captionsetup{type=figure}
    \captionof{figure}{Our method aims at jointly improving geometry and generating texture for an input indoor mesh. The upper figure shows the true room with a panoramic view and a depth map. Then, given a text prompt (in the middle), our model can synthesize new rooms with different styles (in the bottom rows). Note that input mesh is often of low quality, and our method can polish both the texture and geometry. 
    }
    \label{fig:teaser}
\end{center}%
}]

\newcommand\blfootnote[1]{%
  \begingroup
  \renewcommand\thefootnote{}\footnote{#1}%
  \addtocounter{footnote}{-1}%
  \endgroup
}

\begin{abstract}
The techniques for 3D indoor scene capturing are widely used, but the meshes produced leave much to be desired. In this paper, we propose ``RoomDreamer'', which leverages powerful natural language to synthesize a new room with a different style. Unlike existing image synthesis methods, our work addresses the challenge of synthesizing both geometry and texture aligned to the input scene structure and prompt simultaneously. The key insight is that a scene should be treated as a whole, taking into account both scene texture and geometry. The proposed framework consists of two significant components: Geometry Guided Diffusion and Mesh Optimization. Geometry Guided Diffusion for 3D Scene guarantees the consistency of the scene style by applying the 2D prior to the entire scene simultaneously. Mesh Optimization improves the geometry and texture jointly and eliminates the artifacts in the scanned scene. To validate the proposed method, real indoor scenes scanned with smartphones are used for extensive experiments, through which the effectiveness of our method is demonstrated.
\blfootnote{Video results: \href{https://youtu.be/p4xgwj4QJcQ}{https://youtu.be/p4xgwj4QJcQ}.}
\end{abstract}

\section{Introduction}

Commercial depth sensors \cite{zhang2012microsoft} and LiDAR sensors \cite{arkitscenes} on mobile devices have opened a new era in 3D scene capturing for millions of users in their everyday lives. However, the quality of the meshes acquired by these sensors leaves much to be desired, often exhibiting issues such as holes, distorted objects, and blurred textures. In addition, users typically find their surroundings lack variation and may want to further edit and stylize the scene. To solve these problems, this paper demonstrates how to build a 3D scene from text prompts that matches the geometry of a low-quality 3D mesh but differs in style.

Our method is motivated by recent advances in 2D content generation,  especially the diffusion models \cite{song2019generative,ddpm,ddim,rombach2022high,controlnet}. One benefit of diffusion models is to allow user-provided text prompts to guide the image synthesis process, and hence is versatile to generate different styles. 
One straightforward way of extending 2D content generation to a higher dimensional space is treating the 3D scene as a collection of multiview images, and synthesizing the images in a frame-by-frame outpainting manner. However, this approach will suffer from artifacts, and the generated images may not match the geometry of the captured scenes.

Given a 3D scene and a text prompt like "modern style", our work can generate a new scene aligned to the text with coherent geometry and texture.
Our approach involves first generating the 3D scene's texture, followed by the joint optimization of the mesh texture and geometry. We ensure that the generated texture is consistent with the scene's style by starting with a cubemap (a 360$^\circ$ image) at the center of the mesh and then updating the unexplored areas. For the joint optimization of mesh texture and geometry, we propose to identify smooth areas within the generated 2D images and update the mesh geometry accordingly.
\cref{fig:teaser} shows the results of our approach.

Our method differs from previous work in the creation of 3D objects from text \cite{wang2022score,poole2022dreamfusion,lin2023magic3d} and the generation of 3D content based on 2D images \cite{deng2022nerdi,melas2023realfusion,poole2022dreamfusion,instructnerf2023} in two key aspects.
Firstly, we consider a novel and practical setting, as it assumes the presence of a scanned scene, which is common yet largely unexplored. In our distinct setting, we aim to refine existing geometry, as opposed to the previous techniques, which primarily focus on generating new geometry. Secondly, our approach is motivated by a different insight into 2D diffusion models. Our motivation is on the good underlying geometry behind each generated 2D image, whereas previous methods are motivated by generating a set of multi-view 2D images iteratively through diffusion.
Note, we can easily project a mesh-based representation to 2D images, while it is much harder to refine the mesh geometry from 2D inputs. Extensive experiments demonstrate that our approach is accurate and flexible to use in many real applications. 

To sum up, the contributions of this work are three-folded:
\begin{itemize}
    \item We introduce a novel framework that employs 2D diffusion models to edit a given mesh. Our framework facilitates the editing and stylization of both geometry and texture based on textual prompts. 
    \item We design a 2D diffusion scheme for controlling the diffusion models, leading to the production of a scene consistent and structurally aligned texture for the input mesh. 
    \item We conduct extensive experiments using real indoor meshes scanned with smartphones, which verify the effectiveness and reliability of our framework.
\end{itemize}

\section{Related Works}
The domain of 3D content creation  \cite{song2019generative,ddim,ddpm} has significantly improved in recent years. We consider research in this field into two categories. Firstly, using 3D ground truth content for supervision to direct content generation process \cite{chan2022efficient,muller2022diffrf,gu2023nerfdiff}, which is limited due to the availability of high-quality 3D ground truth. The second research category focuses on using the power of existing 2D image generators \cite{rombach2022high} for 3D content creation. Poole \etal \cite{poole2022dreamfusion} proposed Score Distillation Sampling (SDS) to use the structure of the diffusion model, providing supervisory signals to a 3D neural field. Concurrently, Wang \etal \cite{wang2022score} proposed Score Jacobian Chaining to lift pretrained 2D diffusion models for 3D generation. Lin \etal \cite{lin2023magic3d} presented Magic3D, which represents 3D content first through neural fields and then meshes to improve the quality and efficiency of 2D diffusion-guided generation. Fantasia3D \cite{chen2023fantasia3d} proposed to decompose the 3D asset generation as geometry and texture generation problems. SDS has also been applied to convert existing 2D images into 3D models \cite{zhou2022sparsefusion,Xu_2022_neuralLift}. Liu \etal \cite{liu2023zero1to3} proposed adapting existing 2D diffusion models to be camera pose-aware, enabling the direct generation of multi-view images. Chan \etal \cite{chan2023generative} proposes synthesizing a novel view from a single input by incorporating geometry priors with stable diffusion backbones. Another recent work, Text2Room \cite{hollein2023text2room}, uses 2D diffusion models and depth estimation models to generate a textured room mesh from text prompts. 
The biggest difference between our method and the above works is that our method is guided by a scanned mesh, therefore the newly generated 3D scenes will be accurately aligned with the input scene but with different styles.
For editing existing 3D assets, InstructN2N \cite{instructnerf2023} proposed a methodology to update 2D multiview images iteratively. This was based on a 2D image editing model known as InstructP2P \cite{brooks2022instructpix2pix}. InstructN2N performed the editing on 2D images, which meant that the ability to dream entirely new scenes may have been restricted by image-based editing. On the other hand, our approach relies on a geometry-controlled 2D diffusion generation, which implies that it is not hindered by the texture of the original scene.

One big challenge of 3D data collection lies in the imperfect scene scanning results. Because the Lidar on mobile devices is of limited power and resolution, some parts of the scenes are often missed in the point clouds. There have been quite a few researches to improve 3D reconstructions with generative priors, such as self-supervised generation \cite{dai2020sg}, retrieval-based generation \cite{siddiqui2021retrievalfuse}, style transfer \cite{hollein2022stylemesh}. 
Besides the reconstruction problem, some prior research has treated 3D indoor scene generation as an object layout prediction problem \cite{wang2018deep,ritchie2019fast,paschalidou2021atiss}. After predicting the layout, objects are retrieved from a 3D furniture dataset such as 3D-FRONT \cite{3dfront} and placed within the scene. Other approaches, such as Plan2Scene \cite{vidanapathirana2021plan2scene}, use the floorplan and image observations of an indoor scene to predict a textured mesh for the entire scene. Meanwhile, GSN \cite{devries2021unconstrained}, GAUDI \cite{bautista2022gaudi}, and CC3D \cite{bahmani2023cc3d} focus on generating images of indoor scenes through the use of neural radiance fields. 
These research works are inspiring to our work. In practice, we focus on refining indoor scene meshes, especially the smoothness of 3D geometry and the match of geometry and visual textures. We will explain more details in later sections.

\begin{figure*}
\centering
    \includegraphics[width=0.9\textwidth]{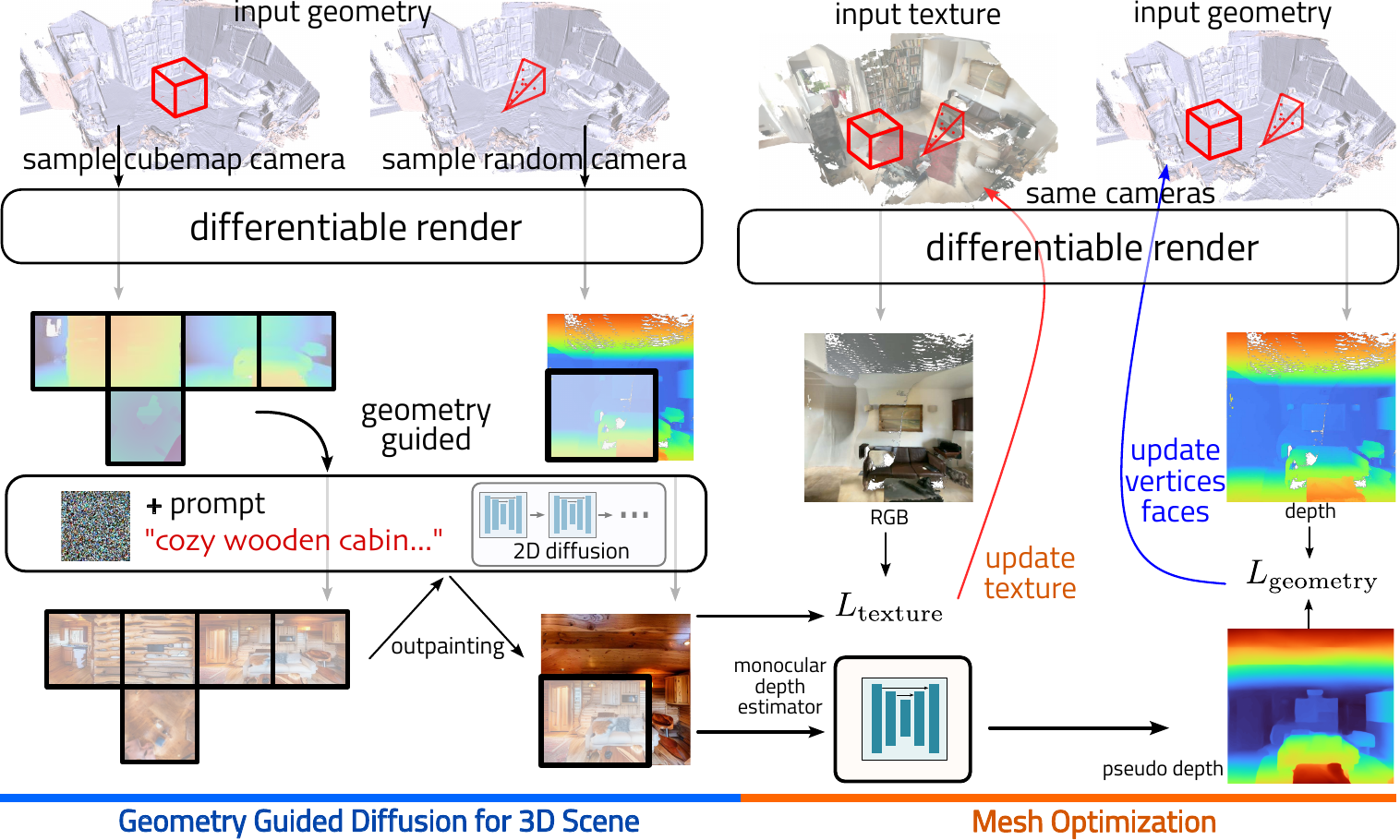}
    \caption{
    The overall framework of our method. Firstly, in the Geometry Guided Diffusion stage for 3D scenes, we create a cubemap representing the scene, followed by outpainting the uncovered areas of the cubemap, as detailed in \cref{subsection-3d-diffusion}. Subsequently, we optimize the mesh texture and geometry. For the geometry optimization, we utilize monocular depth prediction as pseudo supervision and align the smooth areas of the scene, as elaborated in \cref{sec:mesh}.
    }
    \label{fig:framework}
\end{figure*}

\section{Method}
Our approach's input includes a 3D mesh with both geometric and texture information, as well as user-provided text prompts.
Our method is composed of two steps: First, we render the  3D scene to 2D images, and synthesize new styles using a new diffusion process; Then we reconstruct a new 3D mesh with the new textures and polished geometry. An overview of our method is shown in \cref{fig:framework}.

\subsection{Geometry Guided Diffusion for 3D Scene}
\label{subsection-3d-diffusion}

\begin{figure}
    \centering
    \includegraphics[width=\columnwidth]{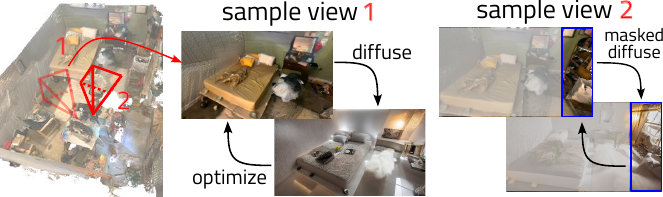}\\
    (a) View-by-view outpainting \\
    \vspace{4pt}
    \includegraphics[width=\columnwidth]{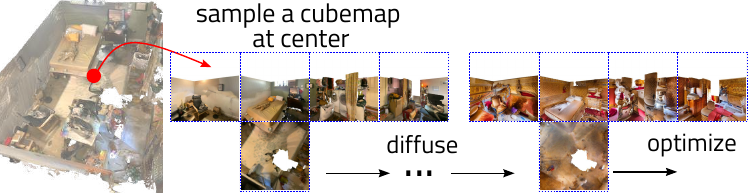}\\
    (b) Cubemap based texture \\
    \caption{
    Methods for generating scene texture. The step ``diffuse'' means generating a 2D image with diffusion models. The ``optimize'' means updating the mesh texture with the 2D generated images (\cf, \cref{eq-update-texture-only}).
    (a) A straightforward baseline based on outpainting with 2D diffusion models. Outpainting is achieved by masked diffusion and the gray area means the masked area remains unchanged through the diffusion. (b) Generating a cubemap for the scene, then optimizing the mesh texture.
    }
    \label{fig:outpainting}
\end{figure}

\begin{figure*}[t]
    \centering
    \includegraphics[width=\textwidth]{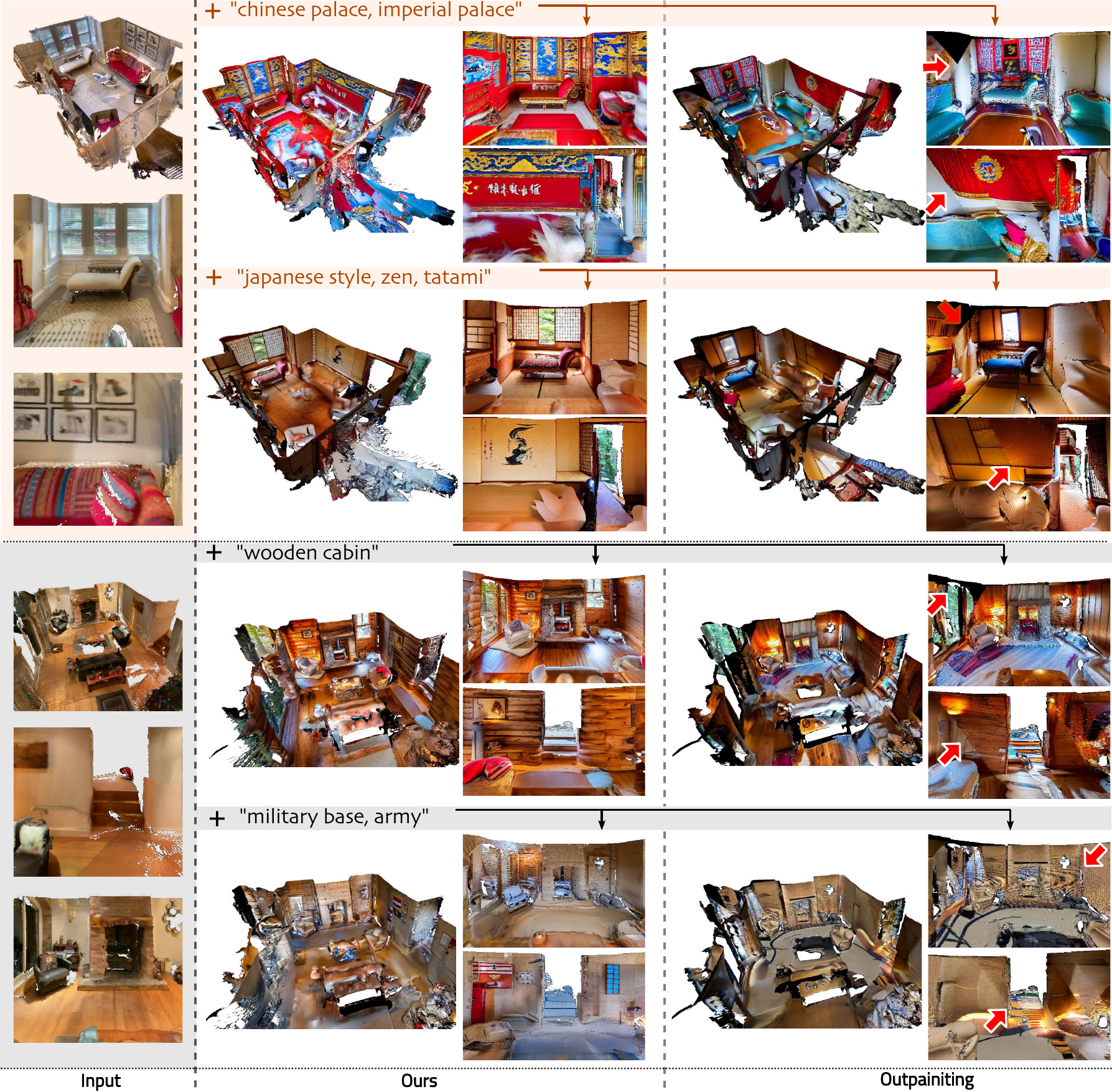}
    \caption{Qualitative comparisons of the output mesh. Outpainting is the baseline method in that textures are outpainted sequentially, while we treat the scene as a whole. Strip shape artifacts can be observed in the outpainting baseline.
    }
    \label{fig:result-outpainting}
\end{figure*}

Synthesizing a new 3D scene is more challenging than synthesizing a 2D image because standard diffusion models  \cite{song2019generative} can easily create inconsistency across different views.  
A straightforward approach for generating scene texture using 2D image diffusion models begins with a randomly positioned camera and iteratively samples neighboring cameras to outpaint \cite{sivic2008creating,yang2019very} the unobserved area, as depicted in \cref{fig:outpainting}. However, we have observed that this baseline method produces noticeable artifacts (\cref{fig:result-outpainting}), which can be attributed in part to the limited outpainting capability of 2D diffusion models.

To avoid the artifacts brought by the view-by-view outpainting generation process, we propose to first generate a 360$^\circ$ image with a central view of the scene. 
A panorama image can be generated with 2D diffusion models by simply extending the diffusion process to cubemap patches \cite{zhange2023diffcollage,bar2023multidiffusion}.
Unlike the classic diffusion model which is conditioned on text prompts, our method is conditioned on both a text prompt $\mathbf{c}_{\mathrm{text}}$ and a depth map $D$, thus the diffusion step is $X_{t-1}=f(X_{t},\mathbf{c}_{\mathrm{text}}, D)$. Following the previous work, we denote the diffusion model as a mapping function denoted as $f:(\mathbb{R}^{H\times W\times 3},\mathcal{C})\rightarrow\mathbb{R}^{H\times W\times 3}$, where $\mathbb{R}^{H\times W\times 3}$ represent the spaces for images with size $H\times W$, and $\mathcal{C}$ represent the spaces of conditional prior including both prompt and image depth.  Further, we denote $X_0=f^{T\rightarrow0}(\mathbf{c}_{\mathrm{text}}, D)$ as the whole diffusion process from random noise and conditioned on the text and depth.

\begin{figure}
    \centering
    \includegraphics[width=1.0\columnwidth]{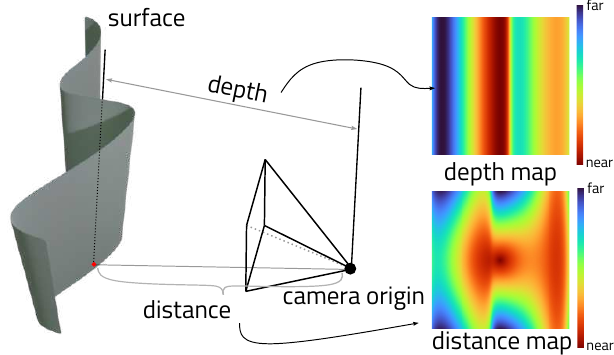}
    \caption{Illustration of the depth map and the distance map. Depth map measures the length between the object plane to the screen plane, while distance map measures the length between points to the camera origin. 
    }
    \label{fig:depth}
\end{figure}

\begin{figure}
    \centering
    \includegraphics[width=\columnwidth]{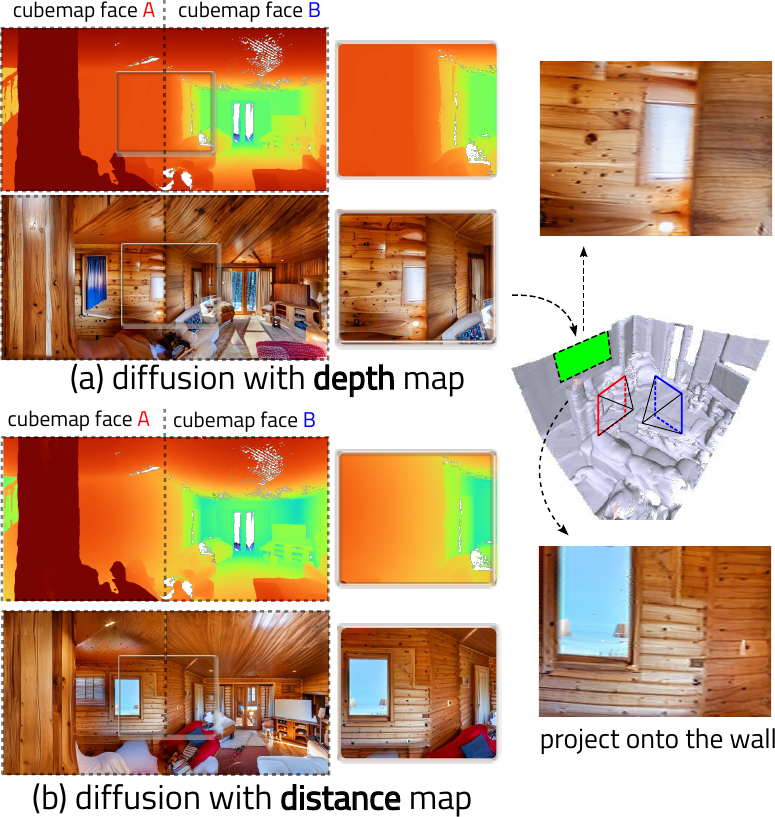}
    \caption{Different controlling effects of the depth map and the distance map. The depth map exhibits rapid change at the joint boundary of the two faces of the cube map. Conversely, the distance map changes smoothly. Generating consistent cube maps with depth control becomes challenging, whereas the employment of distance map results in more consistent texture. However, the distance map results in artifacts such as the window on the wall, as the diffusion model is conditioned on the depth map during training.
    }
    \label{fig:depthexample}
\end{figure}

However, directly using depth map for controlling the generation of cubemap may lead to inconsistency, since the depth value is correlated with camera poses. Different camera poses lead to inconsistency of the depth value in cubemap faces. \cref{fig:depthexample}(a) illustrates the inconsistency of depth map. The depth map associated with each camera is represented in terms of the distance between the camera and the planes. Consequently, the depth values can largely differ from one view to another for the same plane, and lead to artifacts. 
To further reduce the inconsistency, we consider distance map $\hat{D}$ which represents the geometric distance between points and the camera origin. Let a point with world coordinate $\mathbf{p}$ and its associated screen coordinate be $(u,v)$, then the $(u,v)$ pixel on the distance map $\hat{D}$ is $\|\mathbf{p}-\mathbf{o}\|$, where $\mathbf{o}$ is the world coordinate of the camera origin. A comparison between the depth map and the distance map is shown in \cref{fig:depth}.

Distance map $\hat{D}$ and depth map $D$ have different properties in terms of controlling the diffusion process. Structures are well aligned with RGB images in $D$, but distorted in $\hat{D}$. For example, image Laplacian on the planes will be zero in $D$, but not in $\hat{D}$. However, the border in cubemap will with a smoother change with $\hat{D}$, which could be observed in \cref{fig:framework}(b).

To achieve both realistic geometric alignment and border consistency, we propose a blending scheme. For an image patch $p$ at the intersection of cube maps $I_a$ and $I_b$, let $r_a$ and $r_b$ be the ratios of pixels from $I_a$ and $I_b$ in the patch, respectively. We then define $\lambda=|r_a-r_b|$. Each step of the denoising process is calculated using the equation:
\begin{equation}\label{eq:blend}
X_{t-1} = \lambda f(X_t, \mathbf{c}_{\mathrm{text}}, D_p) + (1-\lambda)f(X_t, \mathbf{c}_{\mathrm{text}}, \hat{D}_p),
\end{equation}
where $D_p$ and $\hat{D}_p$ are the depth and distance maps respectively for the patch $p$ being denoised. 
After generating a cubemap, the mesh texture is subsequently updated using a differentiable renderer \cite{Laine2020diffrast}. We then randomly sample cameras in the scene, and the areas not captured by the 360$^\circ$ image are updated through masked generation (\ie, outpainting) with the diffusion model. 
For judging areas captured or not by the cubemap, we can simply project the vertices to the previous cameras and see which vertices are occluded.

\subsection{Mesh Optimization}
\label{sec:mesh}

Both the input and output of RoomDreamer are 3D meshes. We denote 
a mesh as $(V,F,V_c)$, where $V,F,V_c$ are the vertices, faces, and the color of vertices, respectively.  For a specific camera $\pi$, we can render a depth map $D$ and RGB image $X$ at this view:
\begin{align}
X =& \mathcal{R}_X(V,F,V_c, \pi) \\    
D = & \mathcal{R}_D(V,F, \pi)
\end{align}
where $\mathcal{R}$ denotes the rendering function,

In our implementation, we use a differentiable render \cite{Laine2020diffrast}, with which we can back-propagate the gradients to 3D to generate a mesh from synthesized images. Let $\{\pi_k\}$ represent a set of cameras, we can generate a sequence of 
images $\{X^k_0\}$ using the method in 
\ref{subsection-3d-diffusion}. Then we can define a texture-based loss:
\begin{equation}\label{eq:textureloss}
L_{\mathrm{texture}}=\sum_k ||\mathcal{R}_X(V, F, V_c, \pi^k) -X^k_0 ||^2,\\
\end{equation}
Then we get a baseline method of reconstructing the mesh texture by  gradient descent:
\begin{equation}
\begin{aligned}\label{eq-update-texture-only}
V_c &\leftarrow V_c - \gamma   \frac{\partial L_{\mathrm{texture}}}{\partial V_c} \\
&=  V_c - \gamma  \sum_k \frac{\partial L_{\mathrm{texture}}}{\partial \mathcal{R}_X(V,F,V_c, \pi^k)} \frac{\partial \mathcal{R}_X(V,F,V_c, \pi^k)}{\partial V_c} 
\end{aligned}
\end{equation}
Note that \cref{eq-update-texture-only} can only update $V_c$ but not the geometry $V, F$. This is because the differentiable render \cite{Laine2020diffrast} cannot compute the gradient for geometry, i.e.,
$\frac{\partial \mathcal{R}_X}{\partial V}=0$,$\frac{\partial \mathcal{R}_X}{\partial F}=0$. 

A further step is to optimize jointly $V, F$ with $V_c$. Because the input mesh often exhibits low-quality geometry (e.g., with holes), we hope the geometry $V, F$ can be adjusted to match the image sequences $\{X^k_0\}$. An essential observation is that when a synthesized scene contains a smooth region, such as a planar shape, the reconstructed geometry should also be planar. To model this observation, we first reconstruct the depth map $D_k^{\mathrm{gen}}$ from $\{X^k_0\}$ using an off-the-shelf monocular depth estimator $E$ (\eg, MiDaS \cite{Ranftl2022}). Then we define the condition for planar regions
\begin{align}
    |\Delta D_k^{\mathrm{gen}}(u,v)|<\tau
\end{align}
where $\Delta$ is the Laplacian of the depth map $D_k$, and $\tau$ is a threshold for determining smooth areas.

Then, we denote the smooth area as $\mathcal{P}=\{(u,v)|\Delta D_k^{\mathrm{gen}}(u,v)|<\tau\}$. We expect on $\mathcal{P}$, reconstructed depth map $D$ is as smooth as possible, i.e., with Laplacian close to zero. Thus, we get another loss function:
\begin{equation}\label{eq:geometry}
\begin{aligned}
&L_{\mathrm{geometry}}=\sum_k\sum_{(u,v)\in\mathcal{P}} |\Delta D_k(u,v)|,\\
&\text{where}\hspace{1em}  D_k=\mathcal{R}_D(V,F,  \pi^k) 
\end{aligned}
\end{equation}
 Besides the geometry loss, the generated images are used to update the texture of the scene with the following loss.

The overall progress of updating the scene mesh can be represented as follows,
\begin{equation}\label{eq-update-mesh}
\begin{aligned}
    V_c &\leftarrow V_c - \gamma   \frac{\partial L_{\mathrm{texture}}}{\partial V_c}  \\
    V   &\leftarrow V    - \gamma \frac{\partial L_{\mathrm{geometry}}}{\partial V}  \\
    F   &\leftarrow F    - \gamma \frac{\partial L_{\mathrm{geometry}}}{\partial F}.
\end{aligned}
\end{equation}
An algorithm overview is shown in \cref{alg:cap}.

\renewcommand{\algorithmicrequire}{\textbf{Input:}}
\renewcommand{\algorithmicensure}{\textbf{Output:}}
\begin{algorithm}
\caption{Overall pipeline of our method}\label{alg:cap}
\begin{algorithmic}[1]
\Require 
\Statex \underline{System requirements:} 2D diffusion model $f$, monocular depth estimator $E$
\Statex \underline{From user:} mesh $(V,F,V_c)$, text prompt $\mathbf{c}_{\mathrm{text}}$
\Ensure updated mesh $(V^*,F^*,V_c^*)$ with the new style %
\Statex \textit{Step 1: Geometry Guided Diffusion for 3D Scene}
\State set cubemap cameras at scene center %
\State acquire cubemap depth $D$ and distance $\hat{D}$
\State generate cubemap $X_0^{\mathrm{cube}}$ from $f,\mathbf{c}_{\mathrm{text}},D,\hat{D}$ \Comment{use \cref{eq:blend}}
\State sample $K$ random cameras $\{\pi_k\}$ \Comment{for uncovered area}
\For{$k=1$ to $K$}
\If{$\pi_k$ sees areas not covered by $X_0^{\mathrm{cube}}$}
    \State acquire depth $D_k$ from $\pi_k$
    \State generate image $X^k_0$ from $f,\mathbf{c}_{\mathrm{text}},D_k$ 
\EndIf
\EndFor
\Statex \textit{Step 2: Mesh Optimizing}
\For{$X^k_0$ in all generated images}
    \State get $D_k^{\mathrm{gen}}=E(X^k_0)$ using monocular depth estimation 
\EndFor
\For{$n=1$ to $N$}
    \State Update the 3D mesh using \cref{eq-update-mesh}.
\EndFor
\end{algorithmic}
\end{algorithm}

\begin{figure*}[t]
    \centering
    \includegraphics[width=\textwidth]{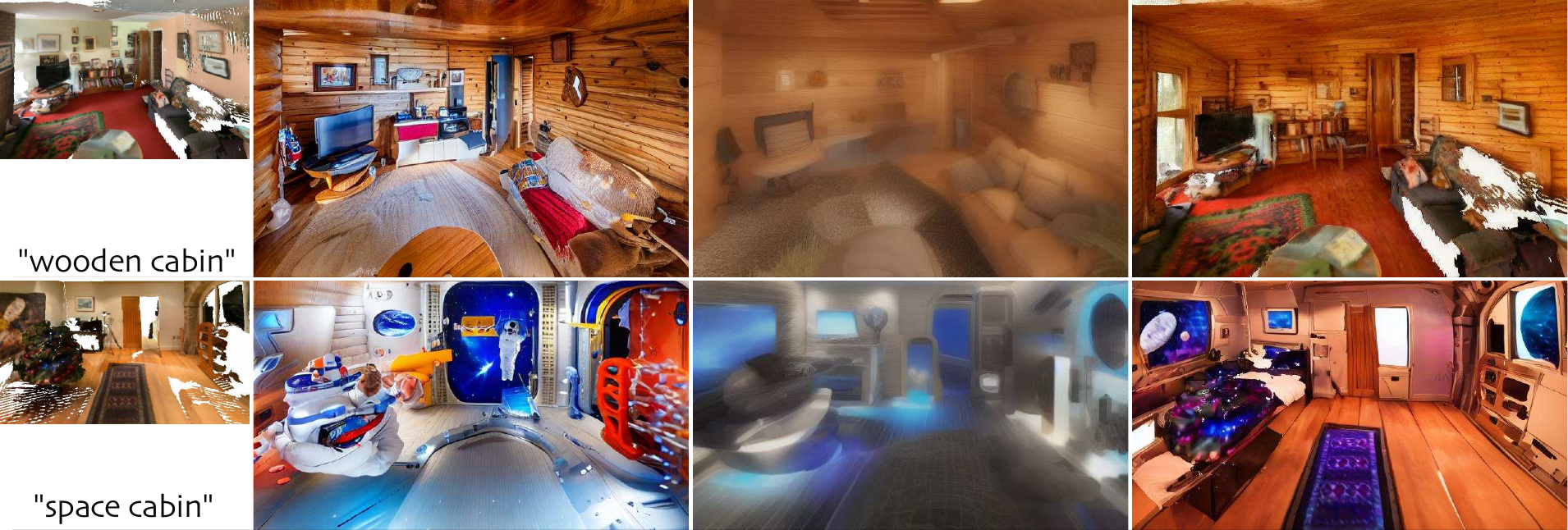}\\
    Inputs\hspace{8em} Ours\hspace{11em} SDS \cite{poole2022dreamfusion}\hspace{9em} InstructN2N \cite{instructnerf2023,brooks2022instructpix2pix}
    \caption{Comparisons with other 2D diffusion based 3D editing methods. SDS \cite{poole2022dreamfusion} can improve the geometry, but the generated texture is kind of blurry. InstructN2N \cite{instructnerf2023} is limited by its backbone InstructP2P \cite{brooks2022instructpix2pix}, which is a purely image-based editing method, and thus may be misled by the presented input image. Our scheme can well handle geometry and texture generation.
    }
    \label{fig:results-sds}
\end{figure*}

\section{Experiments}
Our problem setting assumes inputting with a scanned room mesh, which has not been extensively explored in the existing literature. 
We compare the result of our method with two groups of works: (1) Ablation studies of our own baselines, but without submodules such as geometry-guided diffusion, distance map, smooth region regularization, etc. (2) NeRF-based methods, including Score Distillation Sampling (SDS) \cite{poole2022dreamfusion} and InstructN2N \cite{instructnerf2023}. Since we can project our input mesh to generate multi-view images, which will be used to reconstruct NeRF. Note the reconstruction of NeRF is more computationally expensive, and the reconstructed geometry of NeRF is not always accurate, so we also use the ground truth depth from the mesh input
to boost the reconstruction performance of InstructN2N \cite{instructnerf2023} to get a fair comparison. 

\subsection{Dataset and Implementation Details}
We conducted experiments on the ARKitScenes dataset \cite{arkitscenes}, which comprises real indoor scenes captured by an iPhone. Our method is evaluated qualitatively and quantitatively on the validation set of ARKitScenes, which covers a diverse range of room types and floor plans. 

To generate the cubemap, we set the camera origin at the center of the mesh. For generating depth-based images, we utilize ControlNet \cite{controlnet} and maintain the default hyperparameters, such as the guidance scale and the number of diffusion steps $T$. 
To cover the regions not included in the cubemap, we randomly select $K=100$ cameras around the center and use the masked generation mode of the diffusion model. We predict the monocular depth using MiDaS \cite{Ranftl2022}. During the optimization process, we use the Adam optimizer with a learning rate of 0.001 for optimizing both the geometry $(V,F)$ and the vertices color $V_c$. The optimization is run for a total of $N=1000$ steps. A scene takes around 15 mins to process with one A100 GPU.

\subsection{Comparing with Baselines}
We first conduct a qualitative assessment of our approach in contrast to the outpainting baseline. The results of this analysis are depicted in \cref{fig:result-outpainting}, wherein we compare our cubemap based scene texture generation with the outpainting baseline. Evidently, the scene generated by outpainting exhibits strip-like artifacts that arise from the flawed outcome of the masked generation mode of the diffusion models. Examples of strip-like artifacts can be observed on the walls of the scene. Conversely, our technique consistently produces images with a high-quality and uniform style that is retained throughout the entire scene. This superiority can be attributed to the employment of our cubemap texture generation scheme.

\begin{figure*}
    \centering
    \includegraphics[width=\textwidth]{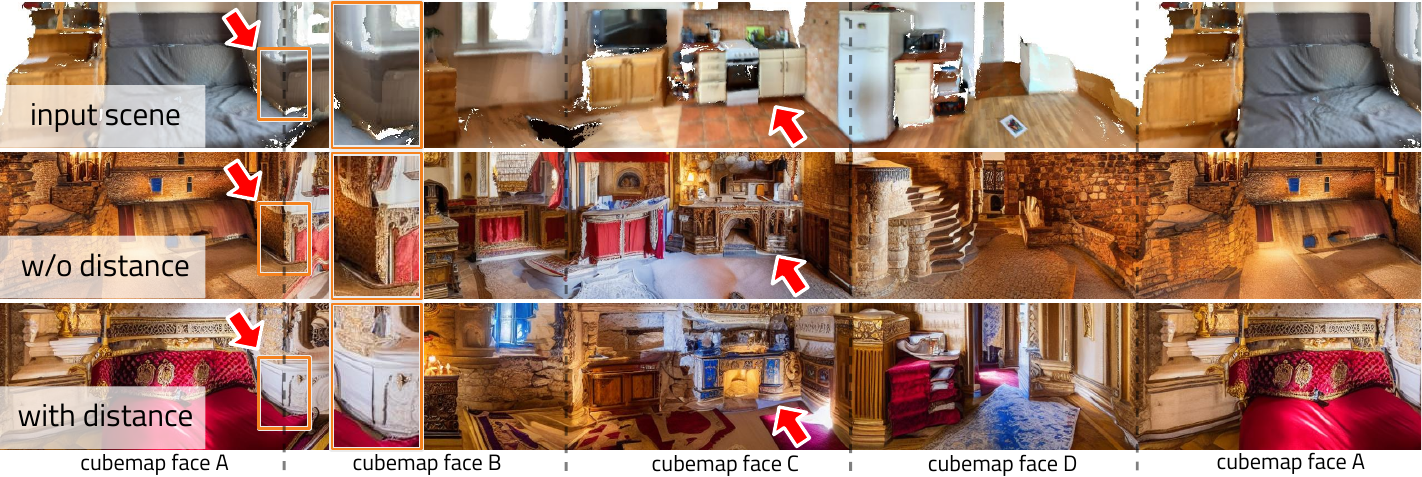}
    \caption{Cubemap generation with and without the distance map blending step (\cf, \cref{eq:blend}). 
    Without distance map blending, the 2D diffusion tends to generate two planes on the border area of the cubemap (\ie, the orange box area). With distance map blending, the border area is treated as one plane.
    }
    \label{fig:results-cubemap}
\end{figure*}

Furthermore, we delve into the impact of blending the distance map (\cf, \cref{eq:blend}) during cubemap diffusion in \cref{fig:results-cubemap}. Notably, a crucial difference can be observed in the region demarcated by the orange box, which represents a patch of the joint area of two cubemap faces. Upon assessing the input scene, we establish that the wall in this area is distorted, implying that the depth signal for diffusion is distorted as well. Specifically, the second row of the figure shows that this area has been treated as a turning corner of two walls instead of a single plane, which indeed is. This generated corner is a result of the distortion in the depth controlling signal. Upon incorporating distance map controlled denoising during the diffusion as shown in the third row, the patch is correctly considered as a single wall plane. Apart from the joint area, we also observe other benefits such as the alignment of areas as indicated by the red arrows.

\begin{figure}
    \centering
    \includegraphics[width=\columnwidth]{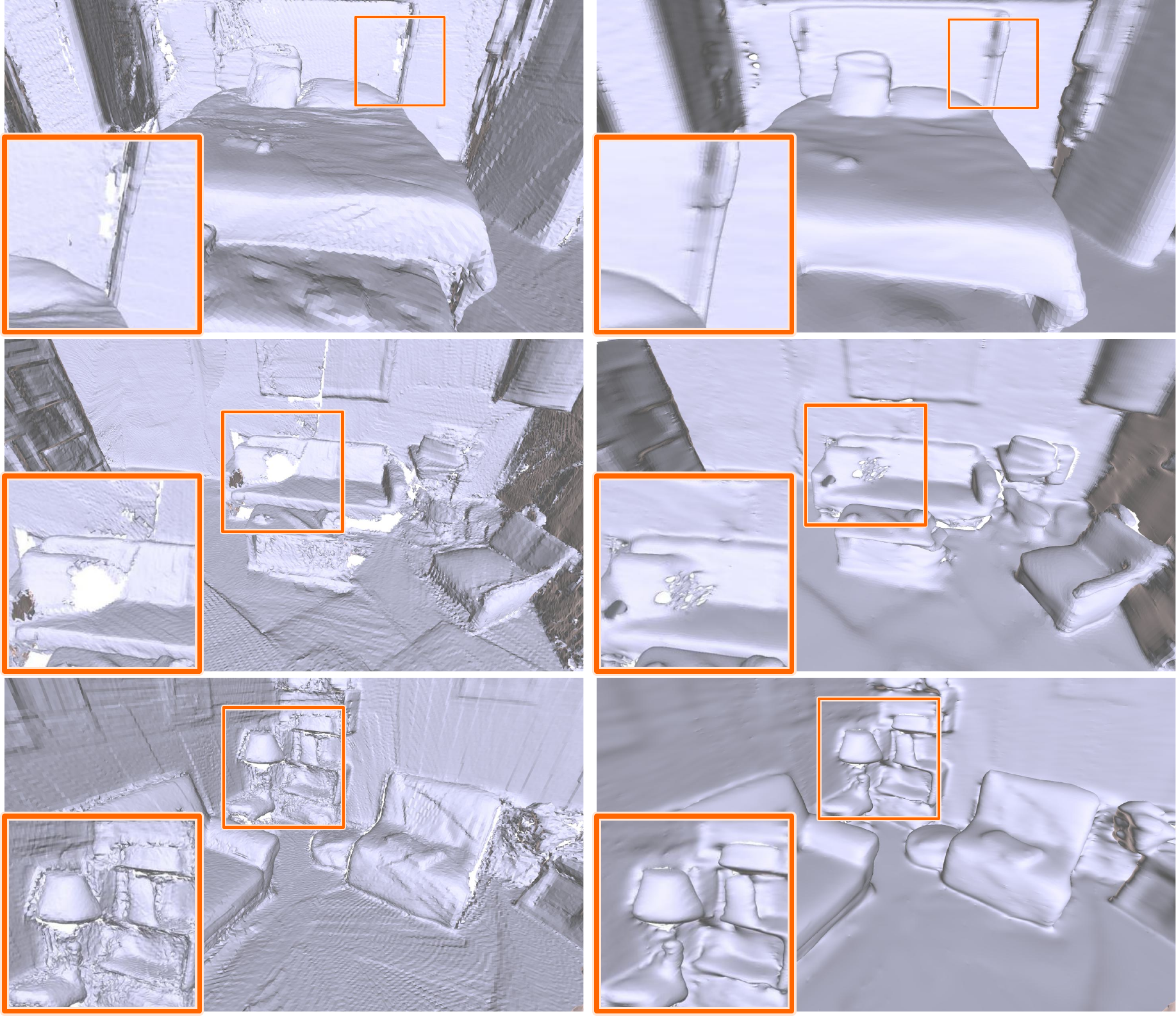}\\
    original geometry \hspace{5em} updated geometry
    \caption{Visualization of the geometry editing. 
    }
    \label{fig:results-geo}
\end{figure}

In \cref{fig:results-geo}, we present a comparison between the original input scene's geometry and the updated scene's geometry. The text prompt used for the generation is ``a royal castle''. It can be observed that after using our method, the mesh is smoother than before. Additionally, our method successfully fills in some holes in the mesh.

\subsection{Comparing with NeRF based Approaches}

In this subsection, we compare our method to two prominent 2D diffusion based 3D generation methods, namely Score Distillation Sampling (SDS) \cite{poole2022dreamfusion} and InstructN2N \cite{instructnerf2023}. 
We used the open-source version \cite{stable-dreamfusion} of SDS with Stable Diffusion as the 2D diffusion model. 

Next, we present a comparative analysis of the performance of our approach versus recently proposed 2D-based 3D editing methods, as depicted in \cref{fig:results-sds}. To ensure a fair comparison, we employ the depth-guided diffusion model (\ie, the same as ours) for our SDS experiments. Upon inspection of the results, we observe that SDS can result in blurred effects, which is similar to the phenomenon recently reported in a study on 2D image generation utilizing SDS \cite{hertz2023delta}. For InstructN2N \cite{brooks2022instructpix2pix,instructnerf2023}, we note that its performance is primarily influenced by the efficacy of its backbone, i.e., InstructP2P \cite{brooks2022instructpix2pix}. It is important to mention that InstructP2P solely relies on the input image and is not conditioned on geometry like depth. Consequently, in InstructN2N, we observe instances where the model is misled by empty regions (\ie, white areas) in the input image, such as the white space on the sofa in the first row. Moreover, employing a purely image-based editing approach may restrict the diversity of the generated images, as demonstrated by the wooden floor in the second row. The comparisons presented in this figure illustrate that our proposed scheme is capable of successfully generating texture with a suboptimal quality input mesh.

\subsection{Quantitative Evaluation}
The quality of a stylized indoor scene is usually subjective, but we have adopted the evaluation approach from \cite{brooks2022instructpix2pix, instructnerf2023} for quantitative analysis of the generated results. This evaluation process is based on the embedding provided by CLIP \cite{Radford2021LearningTV}. There are two metrics used for evaluation. The first metric, called ``Text-Image Similarity,'' computes the inner product between the given text prompt and the generated image. Higher values indicate higher similarity between the text and image vectors, which implies a smaller angle between them. The second metric, referred to as ``Direction Consistency,'' assesses the consistency of the generated scenes across different views. The score is computed as follows: Given two CLIP embeddings of the original input views, denoted as $\mathbf{o}_a$ and $\mathbf{o}_b$, and two CLIP embeddings of the generated views, denoted as $\mathbf{g}_a$ and $\mathbf{g}_b$, the score is calculated as 
\begin{equation}\label{eq:metric}
\frac{(\mathbf{g}_a - \mathbf{o}_a) \cdot (\mathbf{g}_b - \mathbf{o}_b)}{|\mathbf{g}_a - \mathbf{o}_a||\mathbf{g}_b - \mathbf{o}_b|}.
\end{equation}
A lower score implies that the direction of generation is better aligned across different views, indicating greater consistency in the scene generation.
To evaluate the performance, we select a total of 80 meshes from the validation set and create 15 textual prompts. For each mesh, we randomly select 4 different views to test. For calculating Direction Consistency, we use the sampled 4 views and one remaining view, resulting in 4 original-generated pairs. Thus, we obtain a total of 4800 pairs of image-text and original-generated pairs which are used to calculate the evaluation metrics.

We first present a comparison of our method with SDS and InstructN2N in Table \ref{tab:sota}. Our approach outperforms both SDS and InstructN2N with higher similarity and consistency scores. The similarity score of SDS appears to be relatively low, which may be attributed to the blurring effect. The high direction consistency score of SDS reflects the consistent blurring effect. InstructN2N achieves a lower text-image similarity score due to the restriction of its purely image-based editing backbone (InstructP2P). However, the direction consistency score of InstructN2N is good, indicating the effectiveness of its dataset updating scheme. 
\begin{table}
    \centering
    \begin{tabular}{ccc}
    \toprule
        & \begin{tabular}{c}Text-Image\\Similarity\end{tabular} & \begin{tabular}{c}Direction\\Consistency\end{tabular} \\\midrule
       InstructN2N \cite{brooks2022instructpix2pix,instructnerf2023} &  0.2022 & \textbf{0.5416}\\
       SDS \cite{poole2022dreamfusion} & 0.1532 & 0.4184 \\
       \midrule
       Ours & \textbf{0.2543} & 0.5281 \\
    \bottomrule
    \end{tabular}
    \caption{Quantitative comparisons with other editing methods. For the two metrics (\cf, \cref{eq:metric}), a higher value indicates better performance.}
    \label{tab:sota}
\end{table}

\begin{table}
    \centering
    \begin{tabular}{lcc}
    \toprule
        & \begin{tabular}{c}Text-Image\\Similarity\end{tabular} & \begin{tabular}{c}View\\Consistency\end{tabular} \\\midrule
       w/o Cubemap & 0.2123 & 0.5116\\
       w/o Distance & 0.2274 & 0.4678 \\
       w/o Geo Optimize & 0.2017 & 0.5120 \\
       \midrule
       Full Model & 0.2543 & 0.5281 \\
    \bottomrule
    \end{tabular}
    \caption{Quantitative ablation. ``w/o Cubemap'' is the outpainting baseline. ``w/o Distance'' means removing the distance map based blending scheme. ``w/o Geo Edit'' means we do not update the geometry with the pseudo depth supervision (\ie, no $L_{\mathrm{geometry}}$).}
    \label{tab:abla}
\end{table}

We then carry out some ablation studies, as shown in Table \ref{tab:abla}. Our outpainting model exhibits good direction consistency but a low text-image similarity. Removing the distance map blending scheme adversely impacts the text-image similarity score. Furthermore, we find that removing the geometry optimization has a noticeable negative impact on the text-image similarity score. This indicates the necessity of optimizing the geometry when seeking to customize and stylize a scanned mesh.

\section{Conclusion}
In this paper, we tackle the problem of synthesizing a 3D interior scene from text prompts based on a scanned indoor mesh input. We propose a solution that capitalizes on the capabilities of 2D diffusion text-to-image generative models. The primary challenge lies in generating coherent 3D geometry and textural information from the 2D generative priors. To ensure the consistent visual appearance of the whole scene, we first develop a geometry-guided 3D scene texture generation technique. Our key idea is to generate a cubemap of the space, thus achieving a consistent style throughout the different views. We then jointly optimize mesh geometry and texture, based on the pseudo-depth estimated by a monocular depth estimator. Our claimed contributions are validated via experiments on real scanned meshes.

\section*{Acknowledgements}
We are grateful to our colleagues at Apple Inc. for their valuable support in enhancing the quality of this work.

{\small
\bibliographystyle{ieee_fullname}
\bibliography{3deditor}
}

\end{document}